%% file: pipeinfer-conf.tex
\documentclass[conference]{IEEEtran}
\usepackage{fancyhdr}
\usepackage{cite}
\usepackage{amsmath,amssymb,amsfonts}
\usepackage{algorithmic}
\usepackage{graphicx}
\usepackage{textcomp}
\usepackage{xcolor}
\usepackage{xspace}
\usepackage{wrapfig}
\usepackage{makecell}
\usepackage{caption}
\usepackage{subcaption}
\usepackage{float} % Add this in your preamble
\usepackage{booktabs}
\usepackage{url} 

\AtBeginDocument{%
  \providecommand\BibTeX{{%
    Bib\TeX}}}

\def\BibTeX{{\rm B\kern-.05em{\sc i\kern-.025em b}\kern-.08em
    T\kern-.1667em\lower.7ex\hbox{E}\kern-.125emX}}
\begin{document}

\title{PipeInfer: Accelerating LLM Inference using Asynchronous Pipelined Speculation}

\author{\IEEEauthorblockN{\textsuperscript{1}Branden Butler, \textsuperscript{1}Sixing Yu, \textsuperscript{2}Arya Mazaheri, \textsuperscript{1}Ali Jannesari}
\IEEEauthorblockA{\textit{Iowa State University}\textsuperscript{1} \\ \textit{Technical University of Darmstadt}\textsuperscript{2}}
\textit{\{butler1, yusx, jannesar\}@iastate.edu, arya.mazaheri@tu-darmstadt.de}
% \and
% \IEEEauthorblockN{\textsuperscript{1}Sixing Yu}
% \IEEEauthorblockA{\textit{Iowa State University}}
% \and
% \IEEEauthorblockN{\textsuperscript{2}Arya Mazaheri}
% \IEEEauthorblockA{\textit{Technical University of Darmstadt}}
% \and
% \IEEEauthorblockN{\textsuperscript{1}Ali Jannesari}
% \IEEEauthorblockA{\textit{Iowa State University}}
% \IEEEaftertitletext{TEST}
}

\maketitle
\thispagestyle{fancy}
\lhead{}
\rhead{}
\chead{}
\lfoot{\footnotesize{
SC24, November 17-22, 2024, Atlanta, Georgia, USA
\newline 979-8-3503-5291-7/24/\$31.00 \copyright 2024 IEEE}}
\rfoot{}
\cfoot{}
\renewcommand{\headrulewidth}{0pt}
\renewcommand{\footrulewidth}{0pt}

% \author{Branden Butler}
% \email{butler1@iastate.edu}
% \affiliation{%
%   \institution{Iowa State University}
%   \city{Ames}
%   \state{Iowa}
%   \postcode{50011}
% }

% \author{Sixing Yu}
% \affiliation{%
%   \institution{Iowa State University}
%   \city{Ames}
%   \state{Iowa}
%   \postcode{50011}}
% \email{yusx@iastate.edu}

% \author{Arya Mazaheri}
% \affiliation{%
%   \institution{Technical University of Darmstadt}
%   % \city{Ames}
%   % \state{Iowa}
%   \country{Germany}
%   % \postcode{50011}
%   }
% \email{arya.mazaheri@tu-darmstadt.de}

% \author{Ali Jannesari}
% \affiliation{%
%   \institution{Iowa State University}
%   \city{Ames}
%   \state{Iowa}
%   \postcode{50011}}
% \email{jannesar@iastate.edu}

% \renewcommand{\shortauthors}{Butler et al.}

\input{sections/00_abstract}


\maketitle

\input{sections/01_introduction}

\input{sections/02_background}

\input{sections/03_relatedwork}

\input{sections/04_approach}

\input{sections/05_evaluation}

\input{sections/06_conclusion}

% \clearpage

\bibliographystyle{ieeetr}
\bibliography{bib/pipeinfer}

\end{document}

%% file: sections/00_abstract.tex
\begin{abstract}
  Inference of Large Language Models (LLMs) across computer clusters has become a focal point of research in recent times, with many acceleration techniques taking inspiration from
  CPU speculative execution.
  These techniques reduce bottlenecks
  associated with memory bandwidth, but also increase end-to-end latency per inference run, requiring high speculation acceptance rates
  to improve performance. 
  Combined with a variable rate of acceptance
  across tasks, speculative inference techniques can result in
  reduced performance. Additionally, pipeline-parallel designs
  require many user requests to maintain maximum utilization.
  As a remedy, we propose PipeInfer, a pipelined speculative acceleration
  technique to reduce inter-token latency and improve
  system utilization for single-request scenarios while
  also improving tolerance to low speculation acceptance rates
  and low-bandwidth interconnects. PipeInfer
  exhibits up to a 2.15$\times$ improvement in generation speed
  over standard speculative inference.
  PipeInfer achieves its improvement through
  Continuous Asynchronous Speculation
  and Early Inference Cancellation, the former improving latency
  and generation speed by running single-token inference simultaneously
  with several speculative runs, while the latter improves
  speed and latency by skipping the computation of
  invalidated runs, even in the middle of inference.
    
\end{abstract}

%% file: sections/01_introduction.tex
\section{Introduction}

Large Language Models (LLMs) have become wildly popular in
recent times, especially due to their versatility in tasks like language understanding and generation. 
Among the various architectures, decoder-only Transformer models like
OpenAI's GPT series~\cite{brown2020language} and Meta's Llama family~\cite{touvron2023llama} have gained prominence. These models are comprised of a series
of decoder layers, each of which is architecturally identical,
sandwiched between an input embedding layer and an output layer~\cite{touvron2023llama,vaswani2023attention}.
Unlike encoder-decoder models, decoder-only models focus solely on generating output sequences based on the input they receive. During inference, each layer feeds its outputs
directly to the inputs of the successive layer. Their autoregressive nature means that for generating each output token, all layers need to be evaluated iteratively. This design introduces a significant challenge, as the model size exceeds the cache size of the target processor, and a memory bandwidth bottleneck becomes evident~\cite{10.1145/1022594.1022596,shazeer2019fast}. This bottleneck impacts the model's scalability and processing speed, posing challenges in real-time applications.

Despite these challenges, the autoregressive approach is preferred for certain applications due to its ability to maintain high levels of accuracy and contextual understanding in sequence generation. Current research is actively exploring solutions to mitigate these bottlenecks, including advancements in hardware design~\cite{Markidis_2018}, more efficient architectures~\cite{jiang2024mixtral}, and novel parallel processing techniques for exploiting batched inference~\cite{miao2023specinfer}. Batched inference is less susceptible to
bandwidth limitations because of increased temporal and spatial locality, improving processor cache hit rate at the expense of increased latency.

Recently, innovative techniques aimed at mitigating the memory bandwidth bottleneck in LLMs have emerged, drawing inspiration from the concept of CPU speculative execution, including SpecInfer~\cite{miao2023specinfer} and Staged Speculative
Decoding~\cite{spector2023accelerating}. These techniques use
a smaller secondary model to generate a tree of speculative
sequences, that are then batched together and run through
the target model. The key to this process is that inferencing
on a tree of speculations yields probability distributions
for all tokens within the tree, allowing inference to jump
ahead several tokens at a time. The speedup shown by these
techniques is due to the greater efficiency that batched processing
yields.
Building on this concept, other techniques like Medusa~\cite{medusa} and Lookahead Decoding~\cite{fu2023lookahead} have emerged, which adapt the speculative decoding approach by modifying the generation and verification of speculations.

A significant challenge with these speculative techniques is the latency escalation, particularly evident in scenarios of low speculation accuracy.
The crux of the issue lies in the balance between computational time and effectiveness. Although each inference run may take longer, the efficiency is substantially compromised if the speculation accuracy is low, leading to fewer correct and verified tokens. This imbalance results in the computational and time costs outweighing the benefits gained from correct speculations. Additionally, in heterogeneous systems
such as smartphones with CPUs, GPUs, and NPUs, synchronization
costs are too high to make full utilization of all processing elements. Speculation techniques can partially resolve this
by running speculation on one processing element and verification
on another, but current methods only run one phase at a time
and thus still show limited utilization.

In this paper, we rectify these issues by modifying
the speculative inference algorithm to run multiple verification runs
simultaneously, exploiting a pipeline-parallel architecture to
achieve high system utilization while maintaining low communication
overhead. Our design achieves remarkable resilience to
interconnect and computation latency, enabling high-speed inference
on low-cost clusters of disparate commodity hardware. We believe
future work can build on our design to achieve higher utilization
in heterogeneous systems despite interconnect latency and
large throughput differences.
Finally, our design is not bound to the speculative inference algorithms. We believe that
this approach can be extended to other acceleration techniques, such as
Lookahead Decoding or Medusa speculation heads.

Compared to pipeline-parallel speculative inference, we observe
roughly a 1.5-2.15$\times$ improvement in overall generation speed
in our test cases while achieving near-zero slowdown for poor
speculation accuracy. Testing with Gigabit Ethernet as the
interconnect revealed a tolerance to latency and throughput
limitations, increasing its improvement over speculative inference in such scenarios.
For well-aligned models, we observed up to 1.7$\times$
faster generation speed than pipeline-parallel speculation,
and for poorly aligned models, we observed up to a 2.15$\times$
improvement.

We introduce several key contributions to the field of speculative decoding in large language models:
\begin{itemize}
    \item \textbf{Asynchronous Speculation and Inference:}
    We enhanced speculative inference by integrating physically separate compute pipelines to run single-token inference
    or tree verification concurrently with the generation of the speculation tree.
    This modification enables simultaneous processing, significantly improving computational efficiency and reducing latency.
    Time-to-first-token latency reached near-parity with non-speculative iterative inference, while system utilization
    doubled.
    % \arya{stay concrete. provide numbers plz.}
    %We modified SpecInfer
    %to simultaneously run single-token inference and generation of the
    %speculation tree by utilizing physically separate compute pipelines.

    \item \textbf{Continuous Speculation:} By leveraging asynchronous speculation,
    we devised a method of continuously generating speculations in small micro-batches rather than large single batches, improving the end-to-end
    latency and reducing the penalty for low speculation acceptance
    rates. With continuous speculation, we observed that the latency
    reduction was directly proportional to the reduction in batch size. We also
    observed continuous speculation allowed PipeInfer to adapt to low-bandwidth
    scenarios. Continuous speculation improved PipeInfer's generation speed up to 1.5$\times.$
    
    \item \textbf{Pipelined KV Cache Multibuffering:}
    To preserve the causality relationship of the generated tokens, we segmented the KV cache
    sequences into private sections for each speculative run.
    Cache operations are pipelined to maintain coherence during
    inference, allowing speculative runs to avoid
    computation of tokens shared by previous runs, even
    before they have been completed. By exploiting this ability,
    we improved computation throughput by a factor proportional to the alignment of the speculative model.

    \item \textbf{Early Inference Cancellation:} We devised a method
    of flushing invalidated runs from the pipeline
    by back-propagating an asynchronous cancellation signal,
    reducing the performance impact of continuous speculation
    with poorly aligned speculation models. Somewhat counter-intuitively,
    we observed greater speedups up to 2.15$\times$
    for poorly aligned models thanks to early inference cancellation.

\end{itemize}

%% file: sections/02_background.tex
\section{Background and Motivation}
Large Language Models based on the Transformer decoder architecture, including Llama 2~\cite{touvron2023llama} and GPT~\cite{brown2020language},
are built up from a series of decoder layers. Each decoder layer contains an attention module and a multi-layer perceptron~\cite{vaswani2023attention}. The attention module calculates the scores for all tokens in the sequence,
but the key and value vectors are oftentimes cached to prevent needless recomputation.

During inference, each decoder layer is evaluated in sequence, requiring the weights for the attention module and MLP to be
loaded, as well as the relevant entries from the KV cache. The sheer size of most modern models necessitates multiple gigabytes
of memory transfers during a single inference run, resulting in a memory bandwidth bound for small batch sizes~\cite{spector2023accelerating}.
This bandwidth limitation mirrors similar constraints in CPU pipelines, causing significant stalls as the processing elements
wait for values from memory~\cite{10.1145/1022594.1022596}. Inspired by CPU speculative execution, many similar designs for speculative inference were created to alleviate this bottleneck~\cite{spector2023accelerating,miao2023specinfer,medusa}.

% Explain speculation in general, LLM inference procedure, etc.
% Don't explain just specinfer
\subsection{Speculative Inference}
Speculative inference operates through two principal components: the speculation phase and the
verification phase. 

\subsubsection{Speculation} The speculation phase involves a set of secondary models
paired with the primary target model.
The secondary models are chosen to be smaller and faster to run than the primary target model. The
speculative models are first run on the input sequence, generating multiple
output sequences iteratively. 
They persist in this process until the highest output probability drops below a designated threshold that signifies the lowest confidence allowed for a speculative token. Upon achieving this cutoff, the comprehensive tree of speculative sequences, encompassing all potential outputs derived so far, is prepared for the next stage.
% These models are iteratively run until a cutoff probability threshold
% for the most likely token is reached, at which point the entire generated tree of
% sequences is fed into the verification phase.

\subsubsection{Verification} The verification phase takes the entire speculated tree and generates a special
attention mask to ensure that sequences within the tree remain mutually exclusive
in terms of their token visibility. This masking technique
preserves causality relationships inherent in the sequences, while preventing any cross-interference amongst them.
%and ensures that the sequences don't interfere with each other. 

Following this, the speculated tree and its corresponding attention mask are integrated into the target model’s inference pipeline.
When multiple tokens are fed into the inference pipeline, the output
is a set of logit vectors for every token in the input tree. The
verification phase then uses these vectors to iteratively compare the speculated tokens against the
probability distribution of the target model at that token's position. If the speculated
token matches a token that would have been sampled from the distribution, the verification
phase accepts the speculated token and continues verifying other tokens in that sequence.
Conversely, a mismatch leads to sampling a new token from the probability distribution
and then ends the verification.

Running a verification pass is more efficient than running each token through the entire
model. The batch processing approach allows for the reuse of layer weights, reducing the frequency of CPU cache evictions.
Additionally, the speculative inference mechanism includes a set of predictive probabilities for every leaf node in the speculative tree. These probabilities are leveraged to anticipate the next token for sequences that are entirely correct, ensuring that the target model inference is constantly productive, avoiding scenarios where a run is rendered completely pointless.

\subsection{KV Cache}
Upon completion of the verification phase, the system must modify the KV cache for both
the target and the speculative models. The KV cache is a method
of improving generation speed by caching attention vectors for previous
tokens~\cite{vaswani2023attention}. Entries corresponding to rejected tokens
must be removed or otherwise masked in subsequent verification runs.
Popular implementations, such as llama.cpp~\cite{gerganov2023a},
attach metadata to each KV cache cell, identifying the entry's position and which
sequences it belongs to. Such implementations perform the required cache modifications
by modifying the metadata instead. The metadata is then used to construct the attention
mask.

\subsection{Challenges of Speculative Inference}
In standard speculative inference, a notable bottleneck arises from the requirement that the target model's inference must wait until the completion of speculated sequences generated by the speculative models. This waiting period imposes a substantial delay on the time-to-first-token latency, a critical measure of system responsiveness. Additionally,
the complexity and size of the speculative models were limited. Any increase in the inference latency of these speculative models has a direct and proportional effect on the latency of the entire system.

Existing work~\cite{miao2023specinfer} also assumed that the speculative models were run on the same systems
as the target model, requiring either significant GPU VRAM or significant
number of nodes to split the target model weights between. 
At any given moment, only one set of these weights is actively utilized, suggesting the theoretical feasibility of temporarily paging them out to CPU RAM or disk space as needed. However, this approach introduces its own set of challenges and inefficiencies, especially in terms of resource management and access times. The trade-off here lies in balancing the requirement for rapid access to these weights against the limitations imposed by hardware resources, particularly in scenarios where VRAM or computational nodes are at a premium.
We solve this issue by moving the speculative models to a dedicated pipeline, enabling further optimizations
such as asynchronous speculation.

% \arya{here you should conclude the section by saying that we solve this issue by...}
% At any one point, only one set of weights is in use, which would allow paging them out to CPU RAM or disk when needed.

%% file: sections/03_relatedwork.tex
\section{Related Work}
% Move spec infer to here
Speculative decoding has previously been explored through
SpecInfer~\cite{miao2023specinfer} and Staged Speculative Decoding~\cite{spector2023accelerating}.

SpecInfer utilized multiple speculative models executed in parallel with each other,
but did not run the speculative and target models in parallel, resulting in increased end-to-end
latency. Staged Speculative Decoding took a slightly different approach, where
the speculative models were themselves speculatively inferenced, improving overall speed
but incurring even greater latency penalties, as the staged speculations were also not run in parallel
with their respective targets.

Medusa~\cite{medusa} takes a slightly different approach to speculative decoding, adding new sampling heads
to the target model that produce the speculations without a secondary model. Medusa does not
incur significant latency overhead but requires training new sampling heads for the target model.

Lookahead Decoding~\cite{fu2023lookahead} takes a different approach entirely, opting to use Jacobi
iteration to generate multiple tokens simultaneously by generating N-grams based
on the trajectory of the current tokens. The generated n-grams are cached and later verified
in a separate stage. Lookahead Decoding exhibits high utilization and low latency on single-node
systems, but does not account for multi-node systems or systems with slow interconnects.

A recent work named SPEED~\cite{hooper2024speed} uses similar speculative techniques to enhance decoding efficiency, but
instead targets inference on a single GPU. Additionally, SPEED
generates the speculations from the hidden states of previous layers, eliminating the need for a separate speculative model. This approach contrasts with PipeInfer, which utilizes a secondary model for speculation generation. While this difference impacts the computational overhead, it also influences the flexibility and adaptability of the system to different modeling scenarios.
Moreover, invalidation
of speculations results in a pause for the entire system in SPEED,
while PipeInfer continues inferencing all other valid runs at the same time
invalidation and flushing operations are executed. Finally,
SPEED targets specialized models implementing parameter sharing,
while PipeInfer works across a variety of off-the-shelf models.

Exploiting pipeline parallelism to improve inference latency while simultaneously running speculation has been explored previously~\cite{yang2023predictive}. However, this work performed speculation via similar methods to SPEED, predicting a single next
token from the hidden states of intermediate layers, requiring additional sampling heads to be trained. Additionally, this method
only speculates one token in advance at a time, limiting the possible speedup compared to other methods utilizing separate speculative models or heads.

Work has also been done using tensor parallelism, distributing the sub-layer operations across the nodes instead
of whole layers~\cite{tadych2024distributedllama}. However, such methods have been found to suffer from extreme interconnect bandwidth
bottlenecks, even for extremely slow compute nodes like Raspberry Pi 4s. Even as few as eight nodes incurred higher synchronization
time than the time spent running the inference itself.

Non-speculative techniques have also been explored in the area of inference acceleration. A prominent technique is based around
early exit inference, in which some later model layers are skipped if the output from an earlier layer can be used instead.
CALM~\cite{schuster2022confident}, for example, trains a classifier model to detect when an earlier layer is sufficiently confident in its output.
Similarly, Depth-Adaptive Transformers~\cite{elbayad2020depthadaptive} utilize classifiers to predict the depth at which inference can halt, either
per-token or per-sequence. Yet another early-exit style strategy called LITE~\cite{varshney2023accelerating} accelerates inference while maintaining
output accuracy, a problem that plagues other similar systems. Early exit decoding shows promise but requires training classifiers for the target model.

Other non-speculative acceleration techniques focus on modifying the precision at which inference is conducted. Such
techniques include quantization strategies like AWQ~\cite{lin2023awq} and QuIP~\cite{NEURIPS2023_0df38cd1} or pruning
strategies like SparseGPT~\cite{frantar2023sparsegpt} and SparseML~\cite{pmlr-v119-kurtz20a}. Still other techniques such as SqueezeLLM~\cite{kim2024squeezellm}
combine sparsity and quantization. Quantization and pruning approaches generally require an offline conversion step to compress
the original model weights but can be substantially less intensive than a full pre-training run.

PipeInfer does not require any pre-training or conversion steps, but also does not conflict with quantization or pruning techniques.

%% file: sections/04_approach.tex
\section{\NoCaseChange{PipeInfer} Method}
\begin{figure}
    \centering
    \includegraphics[width=1\linewidth]{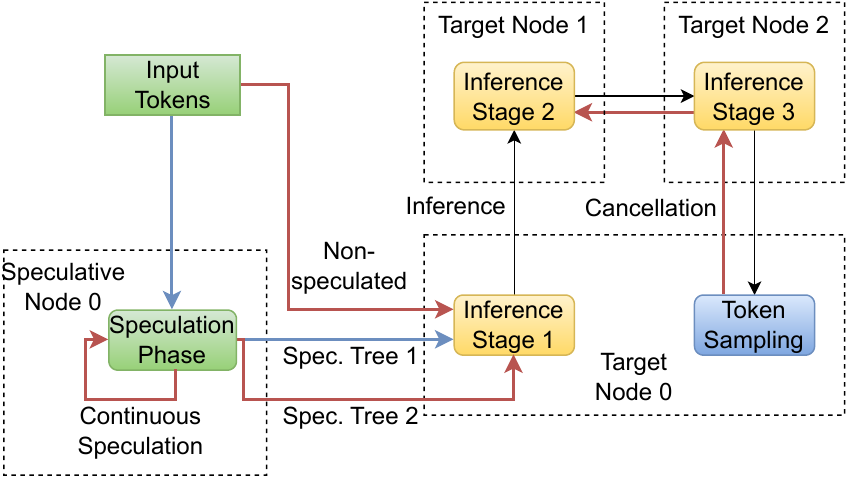}
    \caption{High-level system architecture of PipeInfer. Changes from speculative inference are in red.}
    \label{fig:pipeinfer-architecure}
\end{figure}
Our proposed method, PipeInfer, enhances speculative inference and is composed of four major components:
Asynchronous Speculation, Continuous Speculation,
Pipelined KV Cache Multibuffering, and Early Inference Cancellation.
Our reference implementation is built on top of llama.cpp~\cite{gerganov2023a}, with
inter-node communication accomplished with MPI~\cite{mpi41}.
An overall system diagram is shown in Figure \ref{fig:pipeinfer-architecure}.

\subsection{Asynchronous Speculation}
Requiring the target model inference pipeline to wait for the speculative
sequences to be generated increases both the time-to-first-token (TTFT)
latency and the inter-token latency. PipeInfer reduces these latencies
through Asynchronous Speculation, in which the target pipeline runs
in parallel with speculation.
To accomplish this, PipeInfer utilizes two loosely coupled compute pipelines,
one for the target model and one for the speculative model. After the initial prompt
processing, both pipelines are fed the first generated token. The target pipeline
runs inference on this single token, while the speculative pipeline generates
a tree of speculative sequences.
Once the speculative tree is completed, it is fed into the target pipeline,
which performs verification of the tree.
Upon completion of the first inference run, the logits are transferred to
the head node, which performs sampling, and the process repeats.

For greater speculation accuracy, larger speculative models may be used without
drastically increasing the inference latency of the system because of the
asynchronous design: a larger speculative model requires more time to generate the
speculation tree, but the target pipeline is simultaneously running inference.
Multiple speculative models may also be used, with varying sizes such that the smallest
and the least accurate speculative model quickly generates a speculation tree to keep
the target pipeline full, while the largest speculative model generates a more accurate
tree.

\subsubsection{State tracking}
Each run of the target pipeline is tracked in a data structure containing
the computation graph, the batch used to start the run, and an array of indices mapping
each token in the batch to the corresponding set of logits. The data structure
is created immediately before the run starts and placed in a FIFO queue.
When the run starts, the head node sends configuration data
down the pipeline, detailing information such as the batch size and the array of sequences
per token. The head node then evaluates the first few layers according to
the allocated split. Once finished, the activation tensors are sent
to the next node. All send operations are completed using a buffered implementation, enabling
a sending node to continue before the receiving node is ready.

\begin{figure}
    \centering
    \includegraphics[width=1\linewidth]{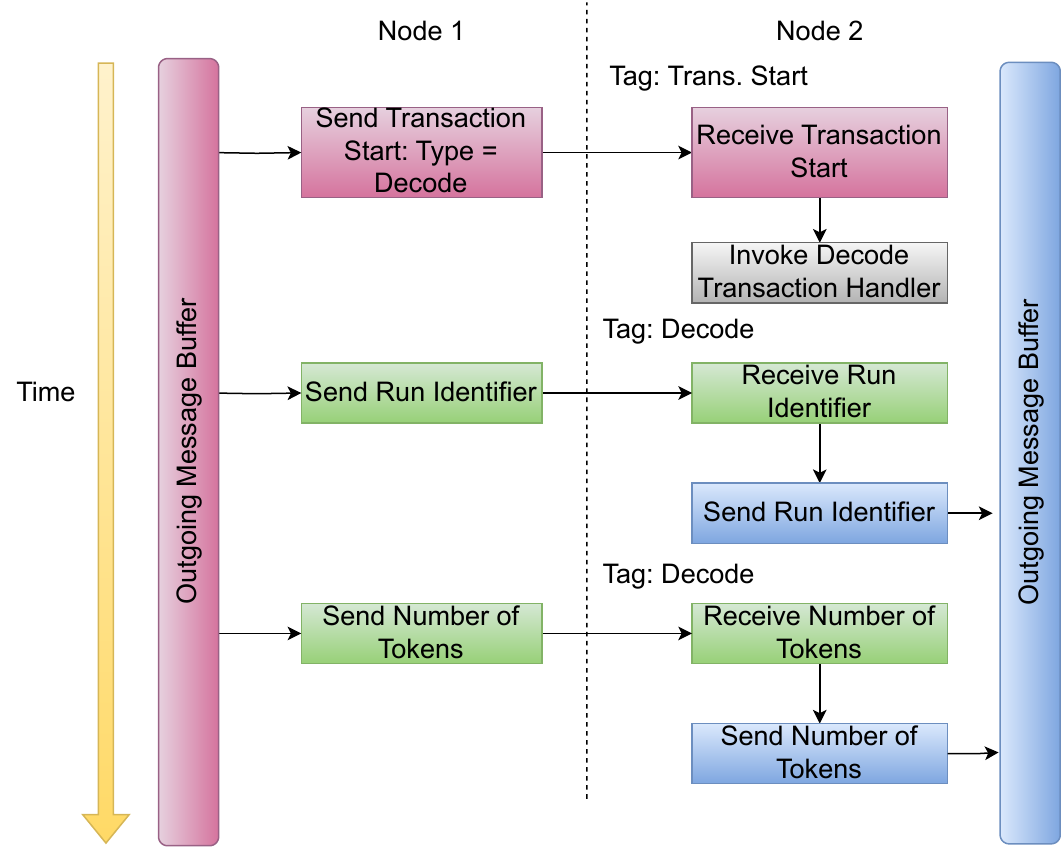}
    \caption{Pipeline communication timeline.}
    \label{fig:pipeline-communication}
\end{figure}

\subsubsection{Pipeline operation transactions}
Most pipeline operations, such as sending configuration data or activation tensors,
are strictly ordered so that activation tensors cannot be received before the requisite
configuration data is received and processed. PipeInfer accomplishes this
using MPI tags: a start message indicates the beginning of a transaction,
a construct defined by PipeInfer to indicate a single atomic operation that must be executed in
the same order as received. The
transaction start message contains the
tag identifying the transaction type, and a handler on each worker node invokes the function
corresponding to that type. All MPI send and receive calls within said function use the tag
attributed to the transaction type. MPI point-to-point communications are non-overtaking
for messages with the same sender, receiver, and tag~\cite{mpi41}, so in this way, we guarantee deterministic
ordering of pipeline transactions. This process is shown in Figure~\ref{fig:pipeline-communication}.

\subsection{Continuous Speculation}
Asynchronous speculation improves end-to-end latency,
but there is still significant under-utilization in the single-request scenario.
After the speculative tree is generated, much of the system remains idle
until the target pipeline completes the initial non-speculative
run. This is not a problem for short pipelines as
the system utilization is proportionally higher, but longer pipelines
suffer from large bubbles of inactivity.

\begin{figure}[ht]
    \centering
    \includegraphics[width=\columnwidth]{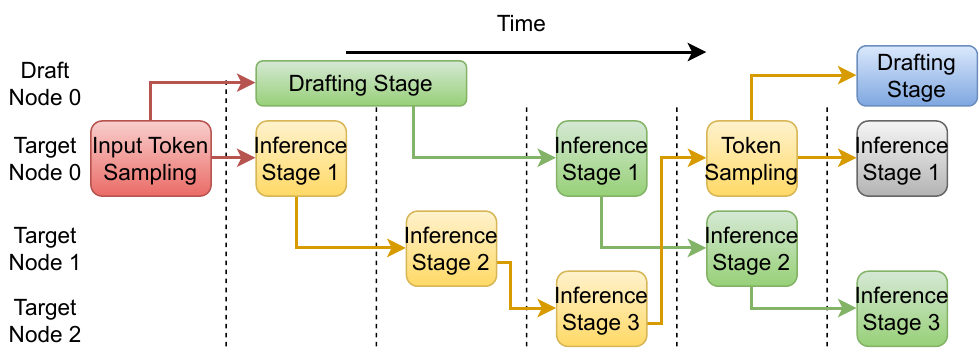}
    \caption{The timeline of PipeInfer using continuous asynchronous speculation.}
    \label{fig:pipeinfer-timeline}
\end{figure}

Continuous Speculation reduces the size of these bubbles
by generating speculative trees whenever the
head node would otherwise be idle. The idle state is determined
by probing for an incoming logits transfer transaction. If a transaction
is waiting to be processed, the head node invokes the
sampling and verification routine. Otherwise, the node generates another
speculation tree. By opportunistically generating speculations,
system utilization improves proportionally to the pipeline depth.
A timeline of continuous speculation is shown in Figure \ref{fig:pipeinfer-timeline}.

Speculative trees generated in this fashion build on the previous trees,
so if an earlier speculation is rejected, many speculative runs are invalidated.

\subsubsection{Microbatching}
An immediate problem with continuous speculation relates to the size of the
speculated trees. In standard speculative inference and asynchronous speculation,
larger trees have the potential to improve generation speed over smaller trees,
as larger batches incur fewer CPU cache misses and less inter-node communication compared to multiple smaller batches.
However, the size of the speculative tree proportionally increases the inference
latency. Additionally, as the depth of the tree increases, the probability of an entire sequence
being accepted decreases due to the divergence between the speculative and target model outputs.
Therefore, larger batches may incur higher latency without improving the number of accepted tokens.

This same principle applies to continuous speculation but is magnified by the greater number
of speculative runs. As a counter-balance, Pipeinfer generates micro-batches of speculations, ranging from 1 to 4 tokens
in size. The smaller batch size improves inference latency at the cost of increased memory bandwidth pressure.
The benefit of micro-batches is three-fold: (1) the imbalance between speculative and non-speculative runs is
reduced, decreasing the size of inactivity bubbles related to this imbalance; (2) splitting
a large speculative batch into many micro-batches allows the system to update the speculative model's
known-correct tokens after verification of only a single micro-batch, improving the overall acceptance rate;
(3) micro-batches enable finer granularity in the context of Early Inference Cancellation,
another component of PipeInfer. Microbatching improves overall inter-token latency, run-to-run latency jitter,
token acceptance rates, and system utilization.

\subsubsection{Reactive speculation}
As the number of speculation trees grows, the probability of all tokens being accepted
drops, as the sequences eventually diverge from the target model. As an attempt to prevent
wasted computation, we added another parameter to continuous speculation called
the confidence cutoff recovery factor. This recovery factor is a floating point value
added to the original speculation confidence cutoff for every successful iteration of
continuous speculation, reset upon acceptance of a completed run. The effect is that of an increasing gradient of required confidence
to continue speculation, reducing the probability of wasted computation.

We also added the inverse, the confidence cutoff decay factor, which is subtracted
from the cutoff threshold when speculation fails and no logits are waiting to be sampled. This decay factor is designed to increase utilization
when waiting for the rest of the pipeline to complete.

With these two factors, PipeInfer's continuous speculation becomes adaptive to
system conditions in real-time, reacting to unexpected slowdowns gracefully
and scaling both up to high-performance university-grade clusters and down
to consumer-grade Beowulf clusters. These factors also allow for PipeInfer to be tuned
towards higher performance or greater power efficiency.

\subsection{Pipelined KV Cache Multibuffering}
Running multiple speculations and a non-speculative inference
simultaneously requires careful management of the KV cache,
the mechanism by which attention-related vectors are cached
to prevent needless recalculation. PipeInfer
uses the KV cache implementation present in llama.cpp~\cite{gerganov2023a}.
The cache metadata enables near-zero
slowdown from copying large numbers of cache cells from one
sequence to another. PipeInfer uses this design
to create multiple partitions of sequence ranges, each range dynamically
allocated on a FIFO policy to a particular inference run.
A queue stores the currently free sequence identifiers,
which designate the beginning of such a range.

\subsubsection{Sequence partitions}

PipeInfer
runs non-speculated inferences using a pre-determined sequence identifier of zero,
termed the canonical sequence, while speculated inferences are allocated a sequence identifier
from the aforementioned FIFO queue.
Combined with the causal attention mask, each inference
run is guaranteed to not interact with cache entries from
other inference runs.

The partitions
act similarly to the back and front buffers in common double-buffering
schemes: while a speculative run is in progress, the partition acts as the
back buffer, only being readable by the assigned inference run.
Once a speculative run is completed, a ``buffer swap'' is performed, where entries
corresponding to accepted tokens are copied to the ``front buffer'', and the partition is
marked as free for use by other runs. The front buffer in this analogy is the canonical
sequence.

\subsubsection{Sequence acceptance and propagation}

When a speculated sequence is accepted during verification,
the cache entries for that sequence are additionally copied to all
other sequences, ensuring that new runs have the correct entries.
Only the entries up until the position of the final accepted token
are copied, ensuring that the cache entries of in-progress runs
are not altered beyond what has already been accepted. Overwriting
existing entries in allocated partitions would cause correctness issues
if the entries were modified before the attention values had been calculated,
potentially causing race conditions and incorrect output. PipeInfer carefully manages operation ordering
to prevent such conditions from occurring.

\subsubsection{Transactions and early cache entry sharing}

Cache operation commands are not broadcast to all nodes simultaneously
but are pipelined like the activation tensors, using the same
transaction mechanism. Doing so ensures
the integrity of in-progress runs and allows
PipeInfer to immediately send a cache copy command after beginning inference
of a non-speculated run, resulting in the guaranteed-correct cache entries
being copied to all sequences immediately after a
node is finished evaluating its set of layers.
Copying these entries enables speculated runs to skip evaluation
of the first token in a sequence and, instead, reuse the cache entry for that token.

\subsection{Early Inference Cancellation}

When running multiple inferences simultaneously, there
is a chance that by accepting multiple tokens in token-tree verification,
some runs in the pipeline become superfluous or invalid. Invalidation
occurs when a speculative run in the pipeline has beginning tokens
that do not match what has been accepted, meaning all tokens in the
run are guaranteed to be rejected. Superfluous runs occur
when all tokens in the run are already accepted; for example,
a non-speculative run may become superfluous if a previous speculative
run had also generated the same token.

\subsubsection{Invalidation detection}

PipeInfer detects
these scenarios through two methods: comparing a run's
starting and maximum ending token positions against the current
accepted tokens' end position, and comparing each run's token
sequences against the currently accepted tokens. Both methods
use a data structure containing the speculated tokens and the
run's maximum and minimum token positions. This data structure is created
when a run is begun and placed in a FIFO queue.
In the former method, if the maximum end position of a run
is less than the current accepted tokens' end position,
then the run is marked as superfluous. In the latter method,
the head node loops over the FIFO queue after every
sampling phase and compares the speculated
tokens against the currently accepted tokens. If the beginning of the speculated
sequence does not match the end of the accepted tokens, then the run is marked
as invalidated.

\subsubsection{Cancellation back-propagation}

Upon detection of superfluous or invalidated runs,
PipeInfer back-propagates a special cancellation signal
through the pipeline. The signal contains
only a uniquely assigned identifier corresponding to the run
that should be canceled. Nodes that have not yet
evaluated the canceled run will skip the evaluation
entirely, improving performance when the pipeline is
saturated or if there is a slower node that would otherwise
become a bottleneck. Nodes currently evaluating a run
probe for a cancellation signal at thread synchronization points,
allowing a node to skip computation even while it is currently processing
the canceled run.

Upon completion of a run, the data structure containing its speculated tokens
and their positions is popped from the FIFO. To maintain consistency,
canceled runs still transfer empty activation tensors down the pipeline, so the ordering
of messages is maintained, and the internal state of each node is kept intact.
Therefore, canceled runs still incur a small amount of communication.

\subsubsection{Performance considerations and conflicts}

Early Inference Cancellation only improves performance when canceling
speculated runs; non-speculated runs are marked as canceled but are still
evaluated in their entirety, and only final sampling is skipped. The reason
behind this difference in behavior is the fact that Pipelined KV Cache Multibuffering
relies on the fact that non-speculated runs are always evaluated in their entirety
to skip the evaluation of the first token in a speculated run. If non-speculated runs
were indeed canceled mid-stream, the subsequent cache-copy commands would copy invalid
entries into the speculated run's sequence partition.

\subsection{Model Accuracy}
PipeInfer accelerates the inference procedure without any loss of model accuracy.
At the sampling and token verification stage, each speculated token is only accepted
if it can be verified that sampling from the target model's output distribution would
have yielded that token. We use the token verification algorithm from SpecInfer \cite{miao2023specinfer}
for this stage.

For the token verification algorithm to guarantee the same model output as in the
non-speculative naive inference mode, the output probability distribution from the model
must also be the same. PipeInfer's careful management of the KV cache guarantees that each
simultaneous speculative inference run is entirely independent, and the use of transactions guarantees
the correct order of operations.

Early inference cancellation only cancels runs that are guaranteed
not to be accepted, and all other runs are allowed to be completed. Non-speculative runs are always allowed
to run to completion, guaranteeing the KV cache is kept in a consistent and valid state.

%% file: sections/05_evaluation.tex
\section{Experiments}

% \input{sections/table/t03_gen_speed}
% % \input{sections/table/t05_latency}
% \input{sections/table/t04_ram}

\subsection{Experiment Setup}
\noindent\textbf{Testbed.} We evaluate the performance of PipeInfer by running various LLMs on various compute nodes with different configurations. The cluster configurations are shown in Table~\ref{tab:cluster_config}. On multi-socket systems, NUMA awareness was enabled,
and the model weights were distributed among the NUMA nodes to take advantage of the independent memory channels.
The 13 heterogeneous nodes comprised five old Dell Optiplexes in combination
with 8 Intel Xeon nodes. This configuration was used to test the resilience of PipeInfer to heterogeneous pipelines,
where slower nodes could stall the pipeline due to greater computation or memory bandwidth bottlenecks.
The Optiplexes were configured with second- and fourth-generation Intel Core i5 and i7 processors, all five of them utilizing dual-channel DDR3 memory.

On cluster C, experiments were executed via a job manager and given exclusive access to all nodes. On clusters A and B,
no job manager was present, and only essential system services were running.

\input{sections/table/t01_model_pairs_test}
\input{sections/table/t02_cluster}

\noindent\textbf{Tested Prompts.} The prompts we tested with were 128 tokens long, formatted according to
the models' expected prompt formats. We used multiple prompts to test different usage scenarios
and to align with each model's expected use case:
\begin{itemize}
    \item The first prompt asked the model to generate a Python program that demonstrates advanced features,
    asking it to withhold any explanation, so the model generates only code.
    \item The second prompt asked the model to write a fictional tale about a warrior named Goliath.
    \item The third prompt used no special formatting and was a randomized excerpt of the Wikitext-2 dataset~\cite{merity2016pointer}.
\end{itemize}

All the models generated 512 output tokens using greedy sampling. We opted to use greedy sampling to maintain the exact generations across all three inference strategies.

% We tested the following models and cluster configurations.
\noindent\textbf{Model Pairs.} The LLM model pairs involved in our experiments are shown in Table~\ref{tab:model_pairs}, including the Llama~\cite{touvron2023llama} and Falcon~\cite{almazrouei2023falcon} model families, as well as a popular Llama merge called Goliath~\cite{alpindale2023}. Goliath is a unique model created by
splicing two Llama 2-70B models together, resulting in a tall and thin model architecture compared to Falcon, which is wider. We added Goliath to our test
cases to determine whether elongated architectures favor one inference strategy disproportionately over others.

\noindent\textbf{Baselines.} We compare PipeInfer with standard,
iterative inference in a pipeline-parallel scenario as well as pipeline-parallel speculative inference, which is an implementation of SpecInfer~\cite{miao2023specinfer}
using a single speculative model.
% Explain why other baselines were not used

\noindent\textbf{Evaluation metrics.} We recorded four primary metrics during our experiments: 

\begin{enumerate}
    \item \textbf{Average generation speed} is measured by recording the total wall clock time between the beginning and completion of inference,
not accounting for initial prompt processing and prefilling.
    
    \item \textbf{Time-to-first-token latency (TTFT)} is measured as the CPU time consumed by the main thread between the completion of the prompt processing phase and the first token acceptance, not including the token sampled at the end of prompt processing. We do not consider the token sampled from the prompt processing stage as the first token because not all inference scenarios require a prompt processing stage, such as when the prompt is cached, and because neither speculative inference nor PipeInfer are engaged during prompt processing.

    \item \textbf{Inter-token latency (ITL)} measures the average time between each accepted token. ITL measurements were taken by recording the CPU time consumed by the main thread in between each accepted token and averaging the recordings. It should be noted that ITL measures the time between each token acceptance and not the time between starting and completing a run.

    \item \textbf{Per-node memory consumption} was recorded through pmap~\cite{pmap}. Model files were memory-mapped, and only the pages that incurred a page fault on the node were included in the memory usage calculations. Before each experiment, we cleared the file cache to guarantee that pages were faulted into the memory attached to the same socket that requested the data, ensuring a consistent experimental environment.
\end{enumerate}

We also observed the output directly and compared it against the single-node inference baseline output to evaluate correctness. This was possible
because the decision to use greedy sampling ensured deterministic output with no run-to-run variance in generated output for a given input prompt.

\subsection{Experiment Results}
For evaluation, we implemented PipeInfer
on top of the popular Llama.cpp framework~\cite{gerganov2023a}. The framework
includes both a reference implementation for speculative inference
and for standard, single-node inference. The framework
also includes an implementation of pipeline-parallel
inference using MPI.

% Add standard deviation and make sure greedy decoding is mentioned

To measure the improvement of PipeInfer, we ran
inference of several models in four scenarios: normal single-node
inference, naïve pipeline parallel inference, pipeline-parallel speculative
inference, and PipeInfer inference. We ran each experiment 10 times and averaged
the results. Unless otherwise specified, all experiments were run on the Intel Xeon Gold
compute nodes.

\input{sections/figtex/small_model_speeds_nova}

\noindent\textbf{Generation speed analysis.}
The Dolphin and TinyLlama pair exhibited acceptance rates of approximately 79\% with the speculative tree size capped at four tokens; the resulting generation
speeds are shown in Figure~\ref{fig:dolphin-speeds-nova}. We observed that PipeInfer improved generation
speed over speculative and iterative inference in all recorded test cases. We also observed that generation
speed using speculative and iterative inference was essentially constant as the number of nodes
increased. We believe this is due to a combination of the extremely fast interconnect and the target model
activation tensors being relatively small.

Switching TinyLlama for Orca 2 7B curiously decreased the overall
acceptance rate to 66\%, decreasing performance for iterative and speculative inference, as shown in Figure~\ref{fig:dolphin-speeds-nova}.
PipeInfer instead shows similar performance to the TinyLlama tests, with slight improvements in the 8 and 32 nodes cases.
We observed that the speculative inference case no longer exhibited constant generation speed as the number of nodes increased; we
theorize that Orca2's confidence in its speculations was high enough to produce larger speculative trees per run, while not aligning
well enough with the target model to substantially reduce the number of runs required, resulting in increased
interconnect bandwidth pressure.

Compared to the Dolphin pairs, the Goliath and XWin-7B pair produced an exceptionally low acceptance rate of 52\%, causing
a decline in the performance of speculative inference as the number of nodes increased. Figure~\ref{fig:goliath-speeds-nova}
plots this decline. PipeInfer's resilience to low alignment is again demonstrated, achieving significantly
higher generation speeds than the other two inference strategies. The highest generation speed was attained at
eight nodes, followed by a slow decline in performance as the number of nodes increased.

To test whether the acceptance rate affects the optimal number of nodes, we replaced XWin-7B with XWin-13B, improving
the acceptance rate to 61\%. Figure \ref{fig:goliath-speeds-nova} reveals that while enhanced alignment increased generation speed, it did not alter the optimal node count, which remained at 8. We did observe
that the 32-node configuration reached parity with the 8-node configuration, while the 15-node configuration
only marginally improved over the XWin-7B test, suggesting non-linearity in the scaling of PipeInfer.

Falcon-180B, paired with Falcon-7B, had a high acceptance rate relative to the size disparity of the models: 68.675\%. Figure \ref{fig:falcon-speeds-nova} reveals that,
with a sufficiently high acceptance rate and sufficiently low speculative model size, speculative inference approaches the performance
of PipeInfer for low numbers of nodes. However, as the number of nodes increases, PipeInfer's performance spikes while speculative inference's
generation speed drops.

Increasing the acceptance rate to 69.47\% by switching Falcon-7B with Falcon-40B reverses the trend:
the difference between the two strategies is greatest at lower numbers of nodes due to the extreme computation requirements of the speculative model.
Figure~\ref{fig:falcon-speeds-nova} shows this trend.

\input{sections/figtex/f09_ttft_nova}

\noindent\textbf{Time-to-first-token latency analysis.}
Examining the time-to-first-token latencies of the previous tests reveals that PipeInfer achieves near-parity with
iterative inference and substantially lower latencies compared to speculative inference. This is shown in Figures~\ref{fig:ttft-dolphin}, ~\ref{fig:ttft-goliath} and ~\ref{fig:ttft-falcon}. We observed that increasing
the speculative model size did not noticeably impact the TTFT latency. PipeInfer, therefore,
becomes an excellent fit for real-time or conversational scenarios. Speculative inference suffered drastically higher
latencies as a consequence of waiting for the speculative trees.

\input{sections/figtex/itl-nova-small}

\noindent\textbf{Inter-token latency analysis.}
Figures~\ref{fig:itl-dolphin}, ~\ref{fig:itl-goliath}, and ~\ref{fig:itl-falcon} show the inter-token latencies
recorded during our experiments. We observed that the ITL measurements followed the trends shown by
the measured generation speed, verifying the correctness of our results.

\input{sections/figtex/efficiency-ablation-speeds-weak}

\noindent\textbf{Memory efficiency analysis.}
Memory usage was recorded during each experiment. We observed that the memory consumption of PipeInfer was equal to that
of speculative inference. Per-node memory usage was reduced for all inference strategies as the number of nodes
increased. Iterative inference maintained lower memory requirements due to the lack of a speculative model.

Comparing the per-node memory usage with the generation speed in Figure~\ref{fig:ram-efficiency} reveals that, of the three inference strategies,
PipeInfer achieves the highest speed-to-memory-consumption ratio, indicating that PipeInfer scales down to low-end
cluster configurations very well.

\noindent\textbf{Constrained hardware performance analysis.}
To measure the effect of computation and bandwidth constraints, we ran several inference
experiments on two more clusters, each using substantially slower hardware and Gigabit Ethernet
interconnects, the results of which are shown in Figure~\ref{fig:low-spec-speeds}. PipeInfer exhibited its greatest speedups over speculative inference in this
scenario, likely due to the increased cost of speculation and the adaptability afforded by
early inference cancellation. We also observed that 
increasing the number of nodes improved performance in all but two cases,
even if the new nodes were slower than the original ones. For the Dolphin
pair, adding additional nodes beyond the 8 Xeon E5 nodes decreased performance only marginally,
while in the Falcon case, performance was reduced substantially. We believe the steep
decline for the Falcon case is a result of the large model size, exacerbating the computation
bottleneck on the slowest nodes. The Goliath test showed improved performance with the slower nodes,
and we believe this is due to the low acceptance rate combined with early inference cancellation and the unique
tall and thin architecture.

An important observation we made is that PipeInfer appears less susceptible to performance degradation
caused by low alignment: PipeInfer's improvement over speculative inference increased for Goliath compared to Dolphin and Falcon.
Conversely, PipeInfer's improvement was marginal when deployed in shallow pipelines.

Testing on these lower-end clusters also showed the extreme latency improvement asynchronous speculation provides.
The time-to-first-token latencies are shown in Figure~\ref{fig:ttft-latency-low-spec}. 
In some cases, PipeInfer achieved lower TTFT latencies than iterative inference; this is
attributed to the fact that one of the nodes is solely dedicated to speculation under PipeInfer, making the
target pipeline one node shorter than in the iterative inference case.

\noindent\textbf{Model output and accuracy.}
We verified that the output of PipeInfer was consistent with the output from standard speculative inference, pipeline-parallel
iterative inference, and single-node inference. Our decision to use greedy sampling resulted in deterministic output for all cases,
and we verified that there was zero deviation between PipeInfer's final output and the output of the other methods.

\input{sections/figtex/ablation}

\noindent\textbf{Ablation Studies.}
We performed several ablation studies with three different model pairs on an 8-node configuration of Cluster C.
The results of the studies are shown in Figure~\ref{fig:ablation}. The baseline of PipeInfer with all features enabled
is included for comparison purposes. Ablating early inference cancellation resulted in decreased generation speed
and increased inter-token latency consistent with our hypotheses. Removing continuous speculation and increasing the
speculative batch size as a counter-balance caused severe performance degradation for the Dolphin and Goliath models
and moderate performance degradation for Falcon. We hypothesize that Falcon's greater resistance to this ablation
can be attributed to early-inference cancellation canceling a significant percentage of continuously speculated
runs. Ablating KV cache multibuffering resulted in incoherent output, and removing asynchronous speculation serialized
all other operations, causing corruption of the KV cache and incoherent output. Therefore, we did not include
performance numbers for these two ablations, since they experienced correctness errors.

\section{GPU Experiments}
We have conducted experiments with PipeInfer using combined GPU and CPU computation.
However, our GPU implementation is based on a later \verb|llama.cpp| commit, after
substantial backend refactoring, and so these results are not directly comparable
with our previous CPU only results. Additionally, the MPI GPU implementation is not yet
fully optimized and we expect continued improvement across the board in the future.

\subsection{Experiment Setup}
\noindent\textbf{Testbed.} Our GPU evaluation utilized a wide variety of hardware to
test the effectiveness of our method across multiple GPU vendors and APIs. The testbench
configuration is shown in Table \ref{tab:gpu_cluster_config}.

We also tested with a large variety of model families, shown in Table \ref{tab:gpu_model_pairs}.

\input{sections/table/t07_gpu_models}
\input{sections/table/t06_gpu_cluster}

\subsection{Experiment Results}

\input{sections/figtex/gpu-speeds}

As shown in Figure~\ref{fig:gpu-speeds}, we observed similar patterns in our GPU results to our previous experiments.
The overall generation speed of PipeInfer was greater than standard speculative inference
in all but one case. However, we observed some significant outliers,
specifically in the experiment involving the Dolphin 2.9 70B and 8B model pair.
This pair is the only Llama 3-based pair we tested, thus we are unable
to conclude whether this outlier is specific to Dolphin 2.9 or whether it is inherent
to Llama 3 models in general.

We also included an additional set of experiments focusing on prompt-to-prompt variance, shown in Figure~\ref{fig:gpu-variance}.
In these experiments, we observed that PipeInfer's overall generation speed remained relatively
consistent across the range of tested prompts, while speculative inference saw more erratic
speed changes.

%% file: sections/table/t01_model_pairs_test.tex
\begin{table*}[ht]

\begin{center}
\caption{List of target models paired with speculation models.}
\label{tab:model_pairs}
\begin{tabular}[tc]{ c | c | c | c || c | c | c | c }
% \hline
%  \multicolumn{4}{|c|}{Model Pairs Tested} \\
 % \hline
\toprule
 Target Model & Size & Quantization & Architecture & Speculative Model & Size & Quantization & Architecture\\ 
\midrule
 Dolphin 2.1~\cite{hartford2023} & 70B & Q3\_K\_M & Llama 2 & \makecell[tc]{TinyLlama OpenOrca~\cite{jeffzhao2023} \\ Orca 2~\cite{mitra2023orca}} & \makecell[tc]{1.1B \\ 7B} & \makecell[tc]{Q4\_K\_M \\ Q4\_K\_M} & \makecell[tc]{TinyLlama \\ Llama 2}\\
 \midrule
 Goliath~\cite{alpindale2023} & 120B & Q2\_K & Llama 2 Merge & \makecell[tc]{XWinLM 0.2~\cite{xwin-lm} \\ XWinLM 0.1~\cite{xwin-lm}} & \makecell[tc]{7B \\ 13B} & \makecell[tc]{Q4\_K\_M \\ Q4\_K\_M} & \makecell[tc]{Llama 2 \\ Llama 2}\\
 \midrule
 Falcon Base~\cite{almazrouei2023falcon} & 180B & Q3\_K\_M & Falcon & \makecell[tc]{Falcon Base~\cite{almazrouei2023falcon} \\  Falcon Base~\cite{almazrouei2023falcon}} & \makecell[tc]{7B \\ 40B} & \makecell[tc]{Q3\_K\_M \\ Q3\_K\_M} & \makecell[tc]{Falcon \\ Falcon}\\
\bottomrule
\end{tabular}
\end{center}

\end{table*}

%% file: sections/table/t02_cluster.tex
\begin{table*}[ht]
    
\begin{center}
\caption{Hardware testbed specifications}
\label{tab:cluster_config}
\begin{tabular}[tc]{ c | c | c | c | c }
% \hline
 % \multicolumn{5}{|c|}{Cluster Configurations Tested} \\
 \toprule
 Cluster Name & Max nodes & CPUs & RAM & Interconnect\\ 
 \toprule
 A & 8 & 2$\times$ Intel Xeon E5-2650 & 128GB 1600 MT/s DDR3 & Gigabit Ethernet \\
 \midrule
 B & 13 & \makecell[tc]{Heterogeneous: \\ 2nd and 4th Gen i5 and i7, 2$\times$ Intel Xeon E5-2650} & 8GB DDR3 & Gigabit Ethernet \\
 \midrule
 C & 32 & 2$\times$ Intel Xeon Gold 6140 & 384GB 2666 MT/s DDR4 & Infiniband EDR 100Gb/s\\
  \bottomrule

\end{tabular}
\end{center}
\end{table*}

%% file: sections/figtex/small_model_speeds_nova.tex
\begin{figure*}
    \centering
    \resizebox{\linewidth}{!}{%
    \begin{subfigure}{.33\linewidth}
        \centering
        \includegraphics[width=\textwidth]{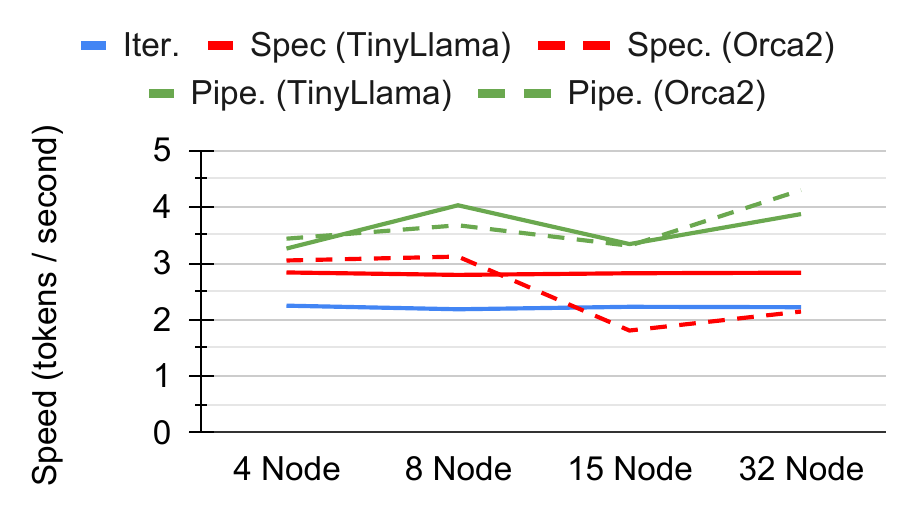}
        \caption{Dolphin-70B speeds using TinyLlama or Orca2 speculative models.}
        \label{fig:dolphin-speeds-nova}
    \end{subfigure}
    \begin{subfigure}{.33\linewidth}
        \centering
        \includegraphics[width=\textwidth]{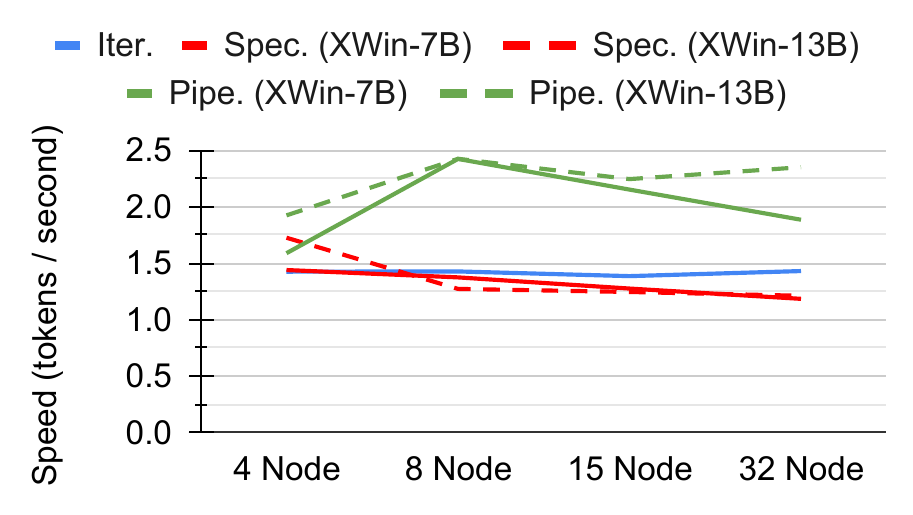}
        \caption{Goliath-120 speeds using XWinLM-7B or XWinLM-13B speculative models.}
        \label{fig:goliath-speeds-nova}
    \end{subfigure}
    \begin{subfigure}{.33\linewidth}
        \centering
        \includegraphics[width=\textwidth]{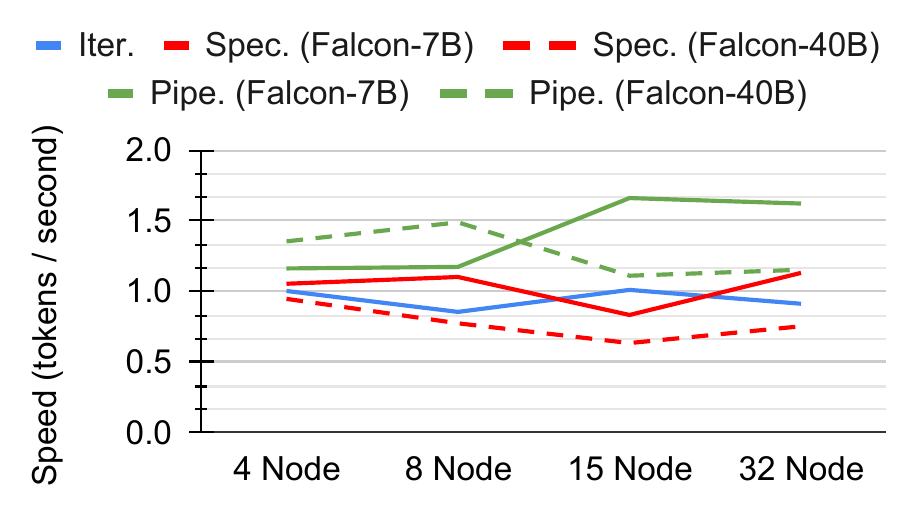}
        \caption{Falcon-180B speeds using Falcon-7B or Falcon-40B speculative models.}
        \label{fig:falcon-speeds-nova}
    \end{subfigure}
  }

  \caption{Generation speeds of model pairs using different inference techniques.
  }
  \label{fig:model-speeds-nova}

\end{figure*}

%% file: sections/figtex/f09_ttft_nova.tex
\begin{figure*}
    \centering
    \resizebox{\linewidth}{!}{%
    \begin{subfigure}{.33\linewidth}
        \centering
        \includegraphics[width=\textwidth]{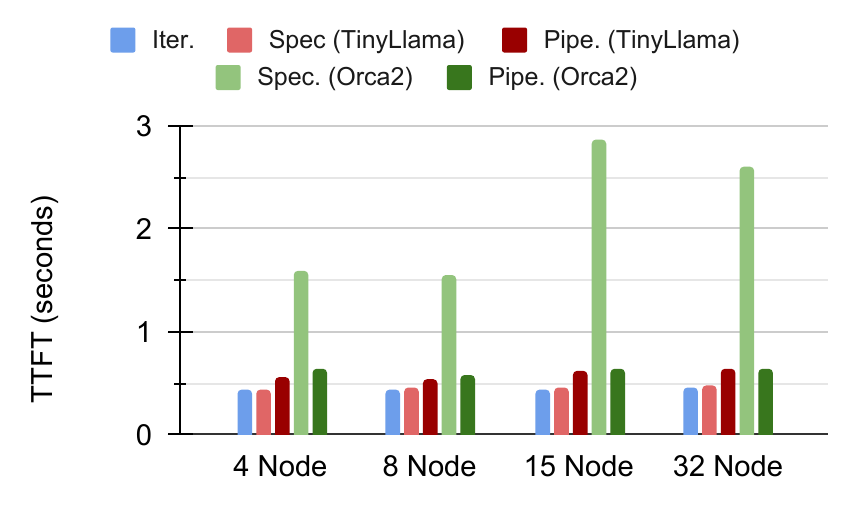}
        \caption{Dolphin-70B paired with TinyLlama and Orca2}
         \label{fig:ttft-dolphin}
    \end{subfigure}
    \begin{subfigure}{.33\linewidth}
        \centering
        \includegraphics[width=\textwidth]{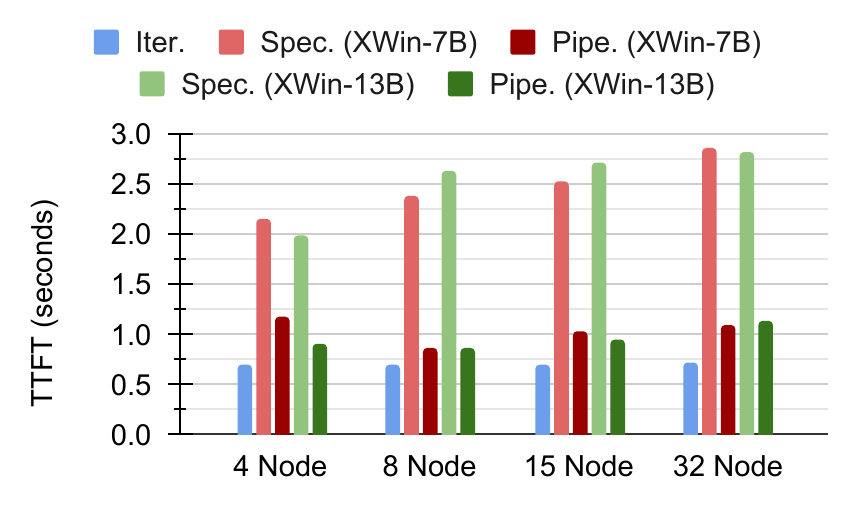}
        \caption{Goliath-120B paired with XWin-7B and XWin-13B.}
         \label{fig:ttft-goliath}
    \end{subfigure}
    \begin{subfigure}{.33\linewidth}
        \centering
        \includegraphics[width=\textwidth]{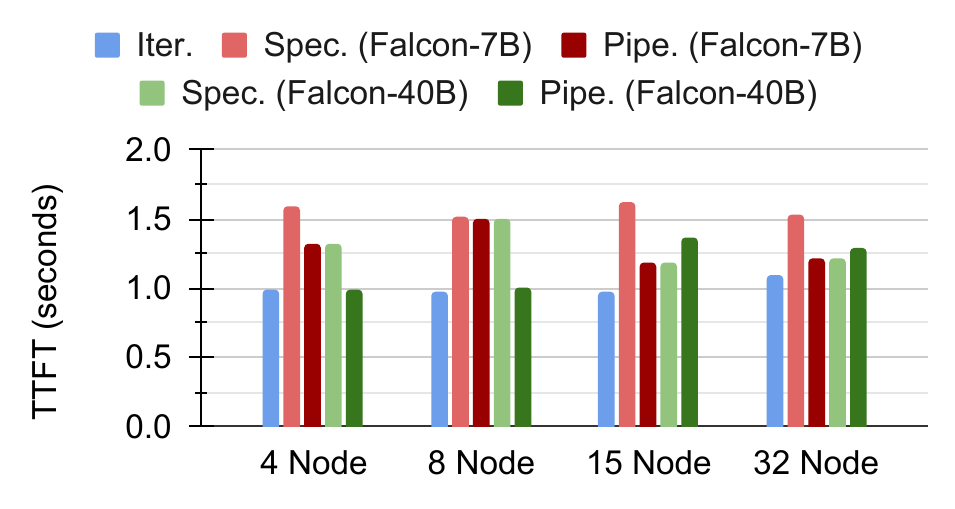}
        \caption{Falcon-180B paired with Falcon-7B and Falcon-40B.}
        \label{fig:ttft-falcon}
    \end{subfigure}
  }

  \caption{Time-to-first-token (TTFT) latencies of model pairs using different speculative models.
  }
\label{fig:ttft-latency-nova-small}

\end{figure*}

%% file: sections/figtex/itl-nova-small.tex
\begin{figure*}
    \centering
    \resizebox{\linewidth}{!}{%
    \begin{subfigure}{.33\linewidth}
        \centering
        \includegraphics[width=\textwidth]{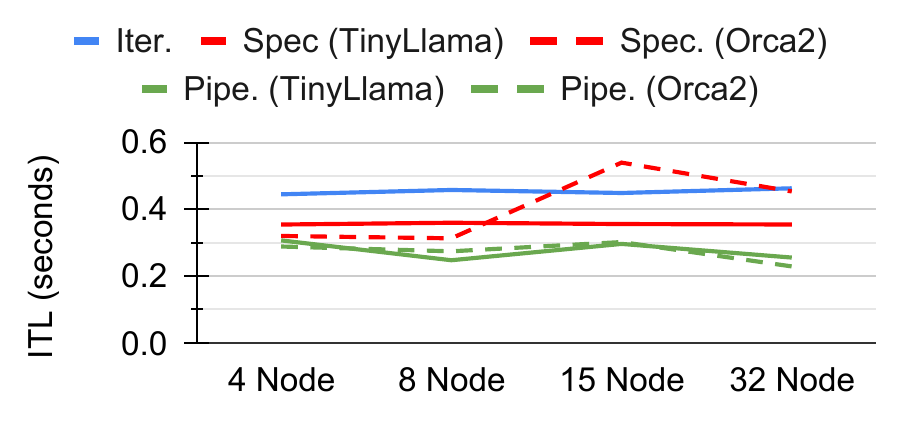}
        \caption{Dolphin-70B paired with TinyLlama or Orca2.}
        \label{fig:itl-dolphin}
    \end{subfigure}
    \begin{subfigure}{.33\linewidth}
        \centering
        \includegraphics[width=\textwidth]{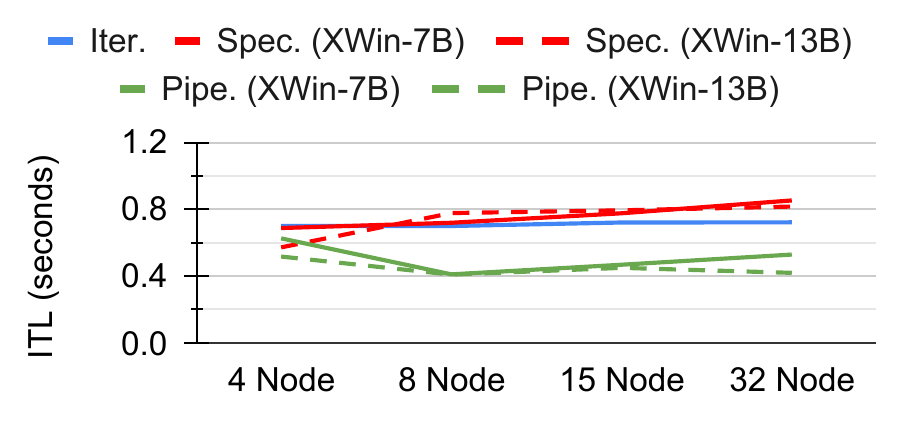}
        \caption{Goliath-120B paired with XWin-7B or XWin-13B}
        \label{fig:itl-goliath}
    \end{subfigure}
    \begin{subfigure}{.33\linewidth}
        \centering
        \includegraphics[width=\textwidth]{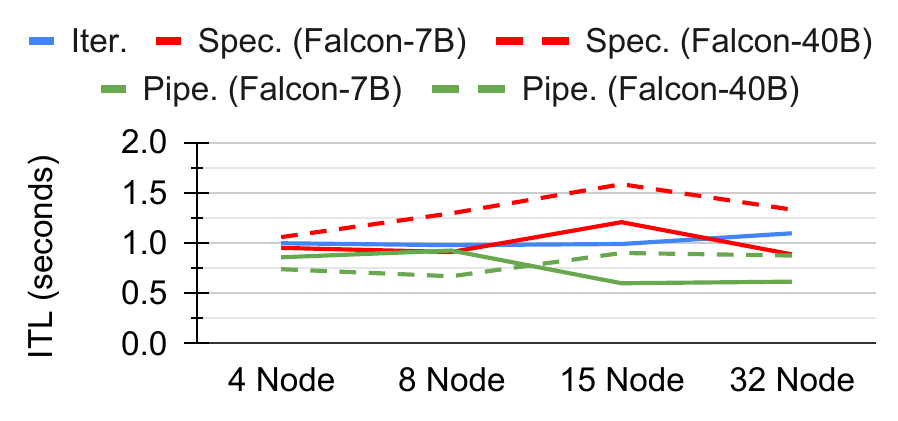}
        \caption{Falcon-180B paired with Falcon-7B or Falcon-40B.}
        \label{fig:itl-falcon}
    \end{subfigure}
  }

  \caption{Inter-token latencies (ITL) of model pairs using different speculative models.
  }
\label{fig:small-model-itl-nova}

\end{figure*}

%% file: sections/figtex/efficiency-ablation-speeds-weak.tex
\begin{figure*}
    \centering
    \resizebox{\linewidth}{!}{%
    \begin{subfigure}{.33\linewidth}
\centering

    \includegraphics[width=\textwidth]{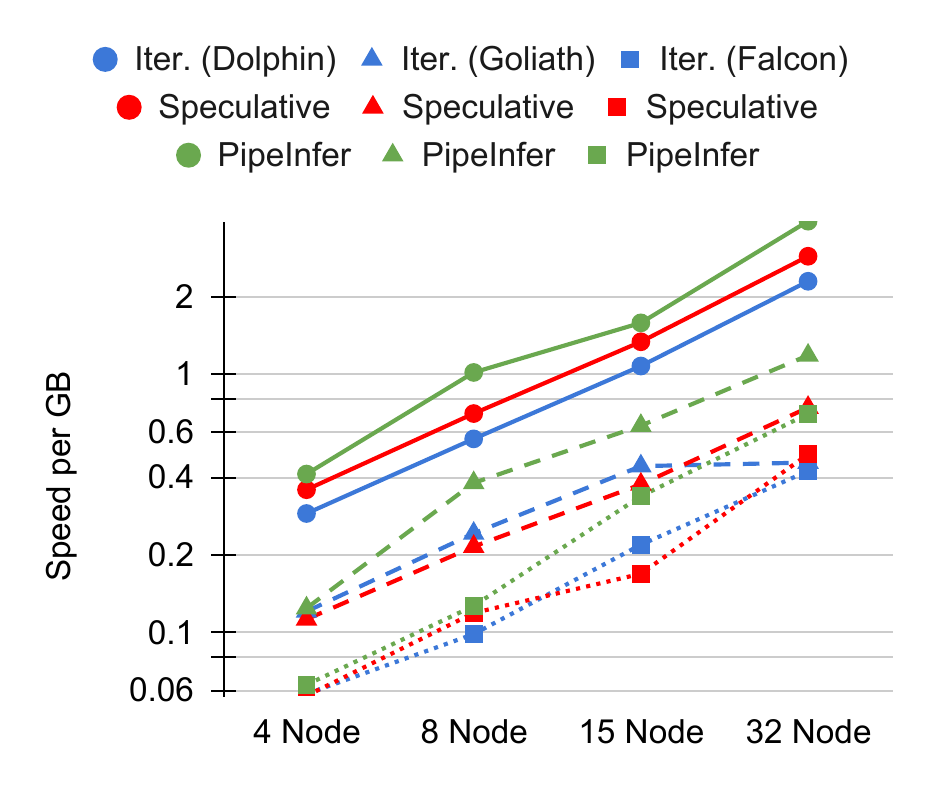}
    \caption{Memory Efficiency. The Y axis is in logarithmic scale.}
    \label{fig:ram-efficiency}
    \end{subfigure}

\begin{subfigure}{.3\linewidth}
\centering
    % \includegraphics[width=\textwidth]{sections/figtex/figures/ablation.pdf}
    % \caption{Ablation studies on 8 nodes with Tinyllama, XWin-7B, and Falcon-7B as speculative models.}
    % \label{fig:ablation}
    \centering
    \includegraphics[width=\columnwidth]{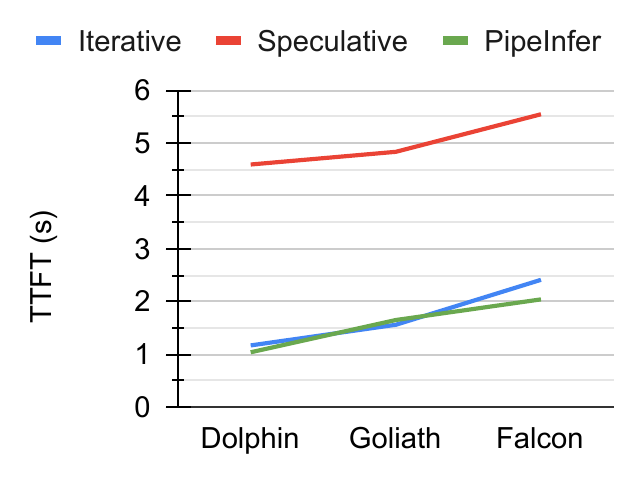}
    \caption{TTFT under different inference methods and models on cluster A.}
    \label{fig:ttft-latency-low-spec}
\end{subfigure}

\begin{subfigure}{.37\linewidth}
\centering

\includegraphics[width=\textwidth]{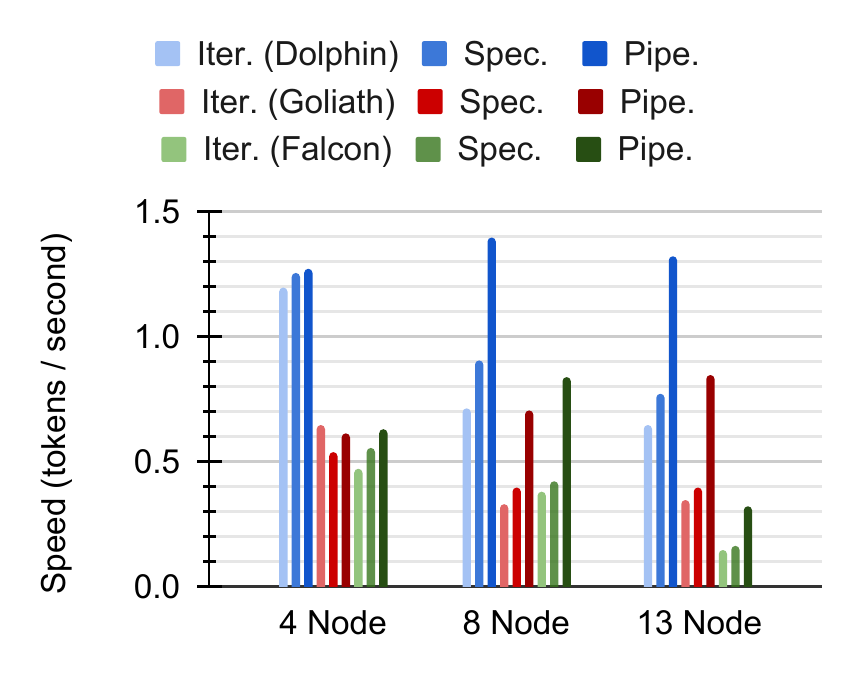}

  \caption{Generation speeds of model pairs using small speculative models on constrained clusters.
  }
  \label{fig:low-spec-speeds}
\end{subfigure}
}
\caption{Resource and performance analysis on constrained hardware.}
\end{figure*}

%% file: sections/figtex/ablation.tex
\begin{figure}
\centering
    \includegraphics[width=\columnwidth,trim={0 0 0 2cm},clip]{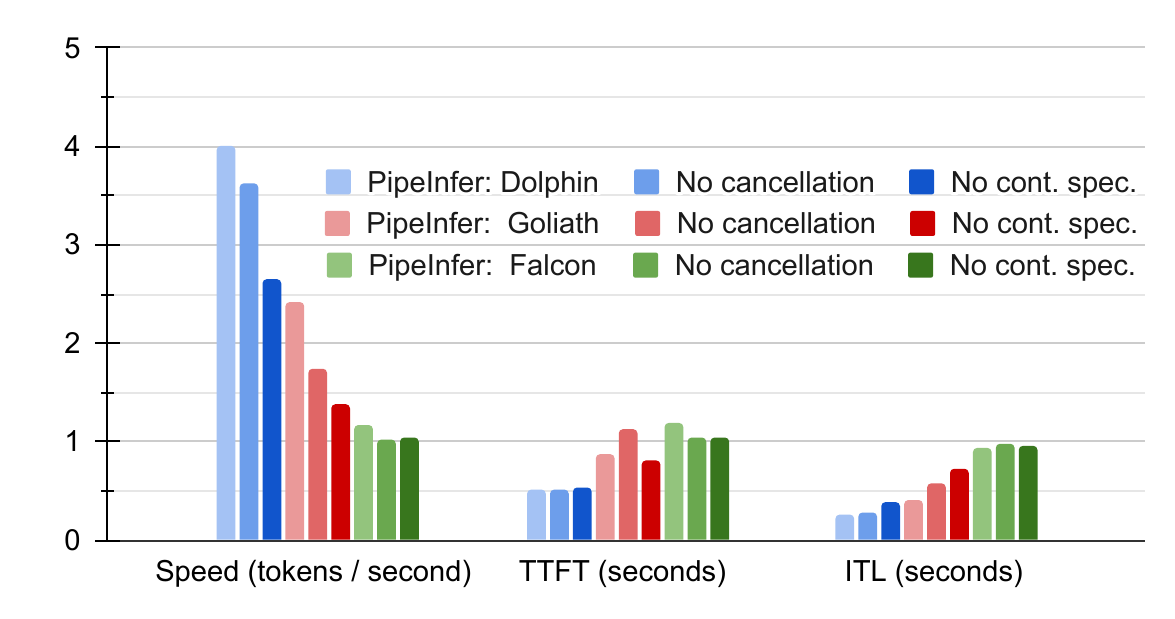}
    \caption{Ablation studies on 8 nodes with Tinyllama, XWin-7B, and Falcon-7B as speculative models.}
    \label{fig:ablation}
\end{figure}

%% file: sections/table/t07_gpu_models.tex
\begin{table*}[ht]

\begin{center}
\caption{List of target models paired with speculation models.}
\label{tab:gpu_model_pairs}
\begin{tabular}[tc]{ c | c | c | c || c | c | c | c }
% \hline
%  \multicolumn{4}{|c|}{Model Pairs Tested} \\
 % \hline
\toprule
 Target Model & Size & Quantization & Architecture & Speculative Model & Size & Quantization & Architecture\\ 
\midrule
 Dolphin 2.1~\cite{hartford2023} & 70B & Q3\_K\_M & Llama 2 & \makecell[tc]{TinyLlama OpenOrca~\cite{jeffzhao2023} \\ Orca 2~\cite{mitra2023orca}} & \makecell[tc]{1.1B \\ 7B} & \makecell[tc]{Q4\_K\_M \\ Q4\_K\_M} & \makecell[tc]{TinyLlama \\ Llama 2}\\
 \midrule
 Senku~\cite{shinoji2024} & 70B & Q3\_K\_M & Llama 2 & \makecell[tc]{TinyLlama OpenOrca~\cite{jeffzhao2023} \\ LongOrca} & \makecell[tc]{1.1B \\ 7B}  & \makecell[tc]{Q4\_K\_M \\ Q4\_K\_M} & \makecell[tc]{TinyLlama \\ Llama 2}\\
 \midrule
 Dolphin 2.9~\cite{hartford2023} & 70B & Q3\_K\_M & Llama 3 & \makecell[tc]{Dolphin 2.9 } & \makecell[tc]{8B} & \makecell[tc]{Q4\_K\_M } & \makecell[tc]{Llama 3}\\
 \midrule
  Qwen~\cite{bai2023qwentechnicalreport} & 33B & Q5\_K & Qwen & \makecell[tc]{Qwen} & \makecell[tc]{7B} & \makecell[tc]{Q5\_K } & \makecell[tc]{Qwen}\\
 \midrule
  Mixtral~\cite{jiang2024mixtral} & 8x22B & Q3\_K\_M & Mixtral & \makecell[tc]{Mistral} & \makecell[tc]{7B} & \makecell[tc]{Q4\_K\_M} & \makecell[tc]{Mistral}\\
 \midrule
  Yi~\cite{ai2024yi} & 34B & Q3\_K\_M & Llama & \makecell[tc]{Yi} & \makecell[tc]{9B} & \makecell[tc]{Q4\_K\_M } & \makecell[tc]{Llama}\\
 \midrule
\end{tabular}
\end{center}

\end{table*}

%% file: sections/table/t06_gpu_cluster.tex
\begin{table*}[ht]
    
\begin{center}
\caption{GPU testbed specifications}
\label{tab:gpu_cluster_config}
\begin{tabular}[tc]{ c | c | c | c | c }
% \hline
 % \multicolumn{5}{|c|}{Cluster Configurations Tested} \\
 \toprule
 Number of nodes & CPUs & RAM & Interconnect & GPUs\\ 
 \toprule
 4 & 2$\times$ Intel Xeon E5-2640 V3 & 128GB 1866 MT/s DDR4 & Infiniband QDR 40Gb/s & \makecell{AMD Instinct MI60, \\ Nvidia Tesla P40, Titan V, RTX 3090} \\
\bottomrule

\end{tabular}
\end{center}
\end{table*}

%% file: sections/figtex/gpu-speeds.tex
\begin{figure}
    \centering
    \includegraphics[width=\columnwidth]{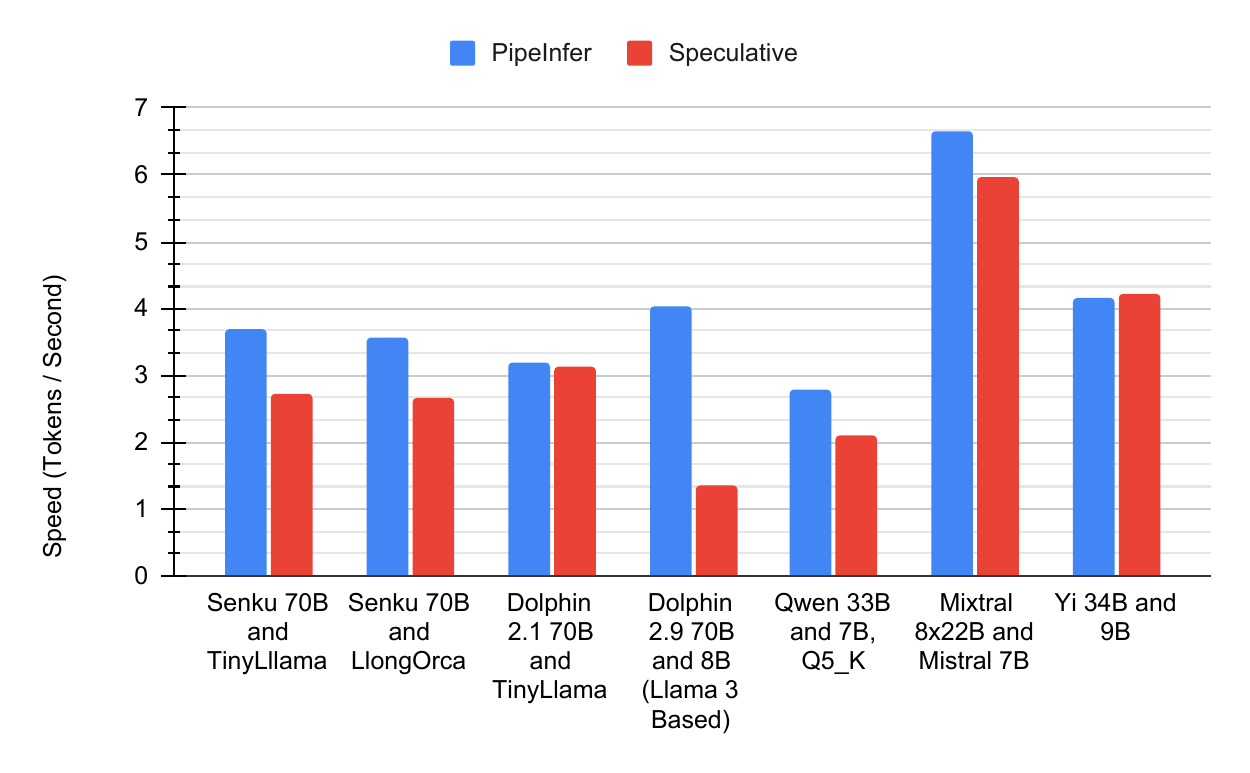}
    \caption{Overall token generation speed on a 4-GPU cluster across many different model pairs.}
    \label{fig:gpu-speeds}
\end{figure}

\begin{figure}
    \centering
    \includegraphics[width=\columnwidth]{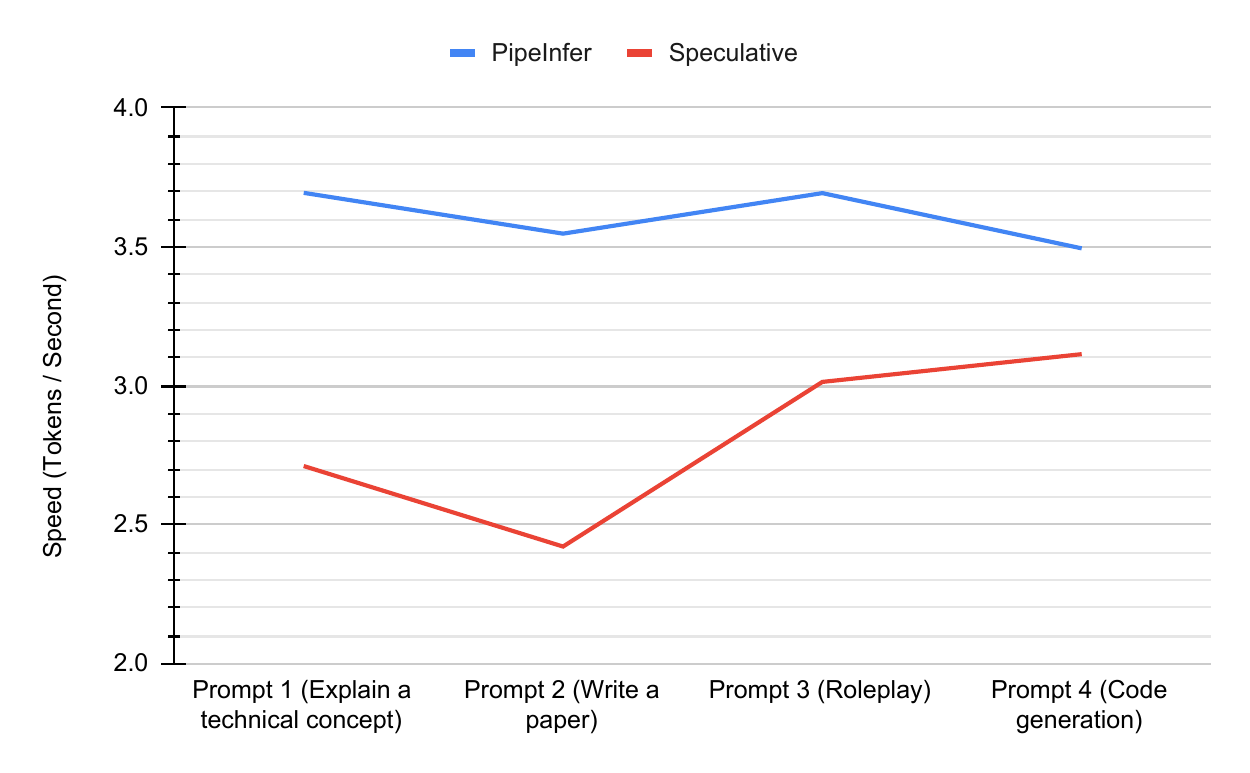}
    \caption{Prompt-to-prompt variance using Senku 70B and TinyLlama 1.1B on a 4-GPU cluster.}
    \label{fig:gpu-variance}
\end{figure}

%% file: sections/06_conclusion.tex
\section{Conclusion and Outlook}

PipeInfer enhances the efficiency and processing speed of LLMs by modifying the speculative inference algorithm to support multiple verification runs within a pipeline-parallel architecture. We achieved not only optimized system utilization and minimized communication overhead but also demonstrated exceptional resilience to latency and computational delays, especially in cost-effective, heterogeneous hardware environments. 
The results show a remarkable 2.15$\times$ improvement in generation speed, maintaining high performance even with low speculation accuracy. Additionally, our approach exhibits robustness in latency and throughput-constrained environments, achieving high CPU and GPU utilization. 

PipeInfer exhibits far-improved performance in multiple scenarios,
and future work may extend it to other inference acceleration strategies,
further improving its performance.
We believe that Lookahead decoding~\cite{fu2023lookahead} and Medusa~\cite{medusa} 
would benefit greatly from PipeInfer
augmentation. We also believe self-speculation techniques like SPEED~\cite{hooper2024speed} or PPD~\cite{yang2023predictive}
could complement PipeInfer's external speculative model approach.

PipeInfer may also be extended to support hybrid parallelization
via multi-GPU nodes, applying tensor parallelism at the local node level and maintaining
pipeline parallelism across the cluster. Alternatively, bandwidth bottlenecks resulting from
the PCIe bus could be alleviated through the application of PipeInfer on a single node.

Bottlenecks in a pipeline can also be mitigated by adding new nodes
in parallel with the slowest nodes, acting as load-balancers. When the
primary node is busy, the secondary nodes can take over the computation of
the designated layers. PipeInfer's use of many independent and simultaneous
inference runs allows such load-balancing without affecting the end results,
so long as the ordering remains consistent.

% Another potential direction for future work could involve either re-enabling
% inference of the initial token in speculated runs, or back-propagating a cache copy
% command to copy from an accepted speculative run to the canonical sequence, overwriting
% invalid data left by a partial inference run. This would allow early inference cancellation
% to also cancel non-speculative runs, potentially improving performance in scenarios
% where speculative model alignment is high.